\documentclass[11pt]{article}
\usepackage{amsmath}

\usepackage[preprint]{acl}

\usepackage{times}
\usepackage{latexsym}

\usepackage[T1]{fontenc}

\usepackage[utf8]{inputenc}

\usepackage{microtype}

\usepackage{inconsolata}

\usepackage{graphicx}
\usepackage{hyperref}
\usepackage{wrapfig}
\usepackage{booktabs}
\title{Adapting Large Language Models to Low-Resource Tibetan: A Two-Stage Continual and Supervised Fine-Tuning Study}

\author{
    Lifeng Chen \\
    University of Georgia
    \And
    Chenyang Zhong \\
    Invictus International School \\
    Hong Kong
    \AND
    Ryan Lai \\
    University of Georgia
    \And
    Tianming Liu \\
    University of Georgia
}

\begin{document}
\maketitle
\begin{abstract}
Adapting large language models (LLMs) to low-resource languages remains a major challenge due to data scarcity and cross-lingual drift. This work presents a two-stage adaptation of Qwen2.5-3B to Tibetan, a morphologically rich and underrepresented language. We employ Continual Pretraining (CPT) to establish Tibetan linguistic grounding, followed by Supervised Fine-Tuning (SFT) for task and translation specialization. Empirical evaluations demonstrate a consistent decrease in perplexity (from 2.98 → 1.54) and substantial improvements in Chinese→Tibetan translation quality (BLEU: 0.046 → 0.261; chrF: 2.2 → 6.6). Layer-wise analysis across 435 layers in Qwen3-4B reveals that adaptation primarily concentrates on embedding and output heads, with mid–late MLP projections encoding domain-specific transformations. Our findings suggest that CPT constructs a Tibetan semantic manifold while SFT sharpens task alignment with minimal representational disruption. This study provides the first quantitative exploration of Tibetan adaptation dynamics for LLMs, and offers an open, reproducible framework for extending multilingual foundation models to low-resource settings. The code of this project can be found on: \href{https://github.com/clf28/Tibetan-Finetuning/tree/main}{https://github.com/clf28/Tibetan-Finetuning/tree/main}
\end{abstract}

\section{Introduction}

\begin{figure}[htbp]
  \centering
  \includegraphics[width=0.45\textwidth]{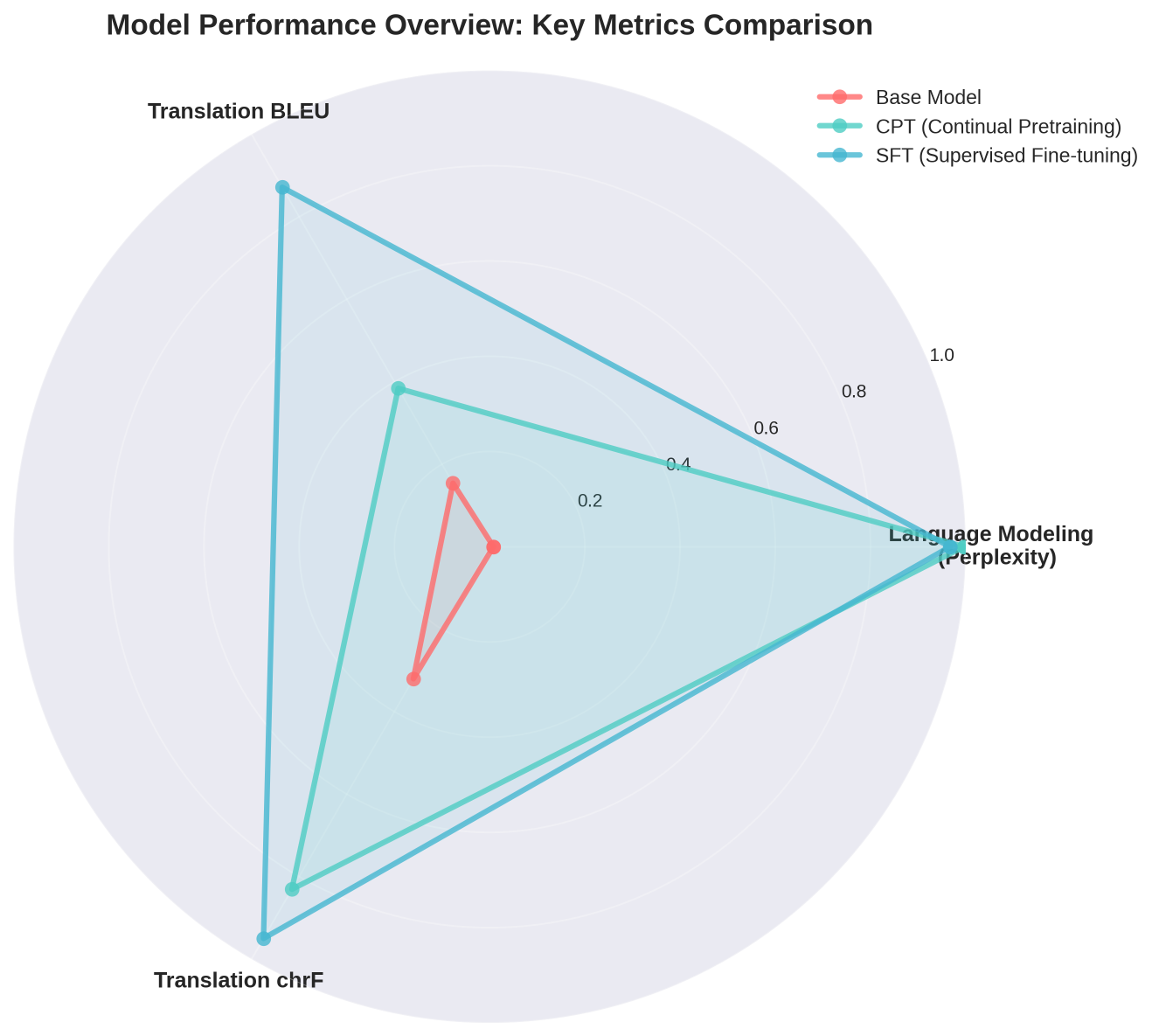}
  \caption{Performance metrics overview showing consistent improvements from Base through CPT to SFT across language modeling and translation tasks.}
  \label{fig:metrics_radar}
\end{figure}

Large language models (LLMs) such as GPT-4 \cite{achiam2023gpt}, LLaMA \cite{touvron2023llama}, and Qwen \cite{bai2023qwentechnicalreport} have achieved remarkable success across a wide range of natural-language understanding and generation tasks. Nevertheless, their linguistic competence remains heavily biased toward high-resource languages such as English and Chinese, while low-resource languages continue to be marginalized \cite{joshi2020state}. Among them, Tibetan is particularly under-served: both large-scale corpora and standardized NLP benchmarks are scarce, limiting the development of inclusive and equitable language technologies for Tibetan-speaking communities.

Tibetan LLMs could also contribute to cultural preservation by supporting translation, search, and educational tools that improve access to literature, historical documents, and recorded oral traditions. Such applications may help younger generations learn and use Tibetan in digital settings, strengthening connections with their heritage. Responsible development should involve Tibetan communities, respect linguistic diversity and cultural ownership, and carefully assess model inaccuracies, enabling these tools to support teachers, scholars, and native speakers in preserving and transmitting cultural knowledge.

Recent advances in multilingual pretraining have attempted to narrow this gap \cite{conneau-etal-2020-unsupervised,costa2022no}. Models such as mBERT \cite{devlin2019bert}, XLM-R \cite{conneau-etal-2020-unsupervised}, mBART \cite{liu2020multilingual}, and NLLB-200 \cite{costa2022no} demonstrate the promise of cross-lingual transfer, yet they still struggle with extremely low-resource languages characterized by limited token coverage, orthographic variation, and complex morphology. In the case of Tibetan, prior work has focused primarily on small-scale translation datasets \cite{10455584,11064426} or monolingual BERT-style models \cite{zhou2023peftt,wang2024tllama}, without systematically examining how modern LLMs can be adapted to this linguistic domain. Moreover, the internal mechanisms of such adaptation—how model parameters evolve across layers when exposed to scarce and structurally unique data—remain largely unexplored.

To address these challenges, we investigate the adaptation of Qwen2.5-3B\cite{hui2024qwen2} to Tibetan through a two-stage training pipeline.
First, a Continual Pretraining (CPT) phase establishes lexical and structural grounding by exposing the model to large-scale Tibetan monolingual text.
Second, a Supervised Fine-Tuning (SFT) phase introduces task-specific and cross-lingual instruction data in Tibetan, Chinese, and English, thereby reinforcing translation and instruction-following capabilities.
This separation of linguistic adaptation and task alignment allows efficient specialization under low-resource conditions while mitigating catastrophic forgetting.

Our empirical evaluation reveals substantial performance gains under low-resource settings. Perplexity consistently decreases from the base model through CPT and SFT, while translation quality improves by over 5× in BLEU and 3× in chrF on Chinese→Tibetan translation. Cross-lingual generalization is also observed in English→Tibetan transfer. Furthermore, a layer-wise weight analysis across 435 layers uncovers that adaptation concentrates primarily in the embedding matrices and mid-to-late MLP gate projections, suggesting a semantic re-anchoring of token and logit spaces rather than a full network re-parameterization. The correlation of CPT and SFT weight changes ($r \approx 1.0$) indicates that fine-tuning consolidates rather than overwrites the linguistic manifold established during continual pretraining. In summary, this study contributes:
\begin{itemize}
    \item A reproducible two-stage framework for adapting large language models to Tibetan, effectively bridging low-resource and multilingual scenarios;
    \item Comprehensive empirical evidence of improvements in both language modeling and translation;
    \item A layer-wise analysis that sheds new light on how large decoder-only architectures internalize low-resource linguistic structure.
\end{itemize}

\section{Related Works}

\subsection{Multilingual Language Models}
Multilingual language models have made significant strides in supporting cross-lingual transfer for diverse languages. Early efforts such as mBERT \cite{devlin2019bert} demonstrated that multilingual pretraining enables zero-shot transfer across languages. XLM-R \cite{conneau-etal-2020-unsupervised} scaled this approach to 100 languages using 2.5TB of CommonCrawl data, achieving substantial improvements on cross-lingual benchmarks, particularly for low-resource languages like Swahili and Urdu. For sequence-to-sequence tasks, mBART \cite{liu2020multilingual} introduced multilingual denoising pretraining, while NLLB-200 \cite{costa2022no} extended translation support to 200 languages through large-scale mining and filtering of parallel corpora.

Despite these advances, extremely low-resource languages such as Tibetan continue to pose significant challenges due to limited token coverage and capacity dilution \cite{conneau-etal-2020-unsupervised}. Recent foundation models in the Qwen series \cite{bai2023qwentechnicalreport,hui2024qwen2} have incorporated broader multilingual support, including Tibetan, through pretraining on diverse web-scale corpora. Qwen2.5-3B, the base model in this study, benefits from native Tibetan tokenization, which provides a more favorable starting point for task-specific adaptation compared to models with limited or no Tibetan coverage.

\subsection{Tibetan NLP}
Tibetan natural language processing remains in its early stages due to data scarcity, orthographic variation, and morphological complexity. Recent surveys \cite{huang2024tibetan} highlight the fragmented landscape of Tibetan resources, with most datasets limited to small-scale parallel corpora for machine translation \cite{10455584,11064426}. Prior translation efforts have focused primarily on Tibetan-Chinese language pairs using recurrent architectures such as GRUs \cite{10455584}, though these approaches lack the generalization capacity of modern LLMs.

Recent efforts have introduced Tibetan-specific language models. Zhou et al. \cite{zhou2023peftt} explored parameter-efficient fine-tuning (PEFT) techniques such as prompt-tuning and adapters for Tibetan pretrained models, demonstrating improvements in downstream tasks with minimal parameter updates. Wang et al. \cite{wang2024tllama} developed T-LLaMA, a Tibetan adaptation of LLaMA2 through continual pretraining, marking the first large-scale decoder-only model for Tibetan. Evaluation benchmarks such as TLUE \cite{wang2025tlue} have also emerged to standardize assessment of Tibetan language understanding.

However, prior work has not systematically examined the two-stage pipeline of continual pretraining followed by supervised fine-tuning for Tibetan LLMs, nor has it investigated the internal parameter changes during adaptation. Our study addresses this gap by leveraging Qwen2.5-3B's native Tibetan support, which substantially reduces the vocabulary adaptation overhead and facilitates more effective fine-tuning.

\subsection{LLM Fine-Tuning Techniques}
Continual pretraining and supervised fine-tuning have become essential paradigms for adapting large language models to specific domains and tasks. Gururangan et al. \cite{gururangan2020dont} demonstrated that domain-adaptive pretraining (DAPT) and task-adaptive pretraining (TAPT) yield significant performance gains by exposing models to in-domain text before task-specific fine-tuning, particularly under low-resource conditions. This two-stage approach mitigates catastrophic forgetting by separating linguistic adaptation from task alignment.

For instruction-following capabilities, supervised fine-tuning on instruction-formatted datasets has proven highly effective. Alpaca \cite{taori2023alpaca} and Vicuna \cite{chiang2023vicuna} popularized instruction tuning by fine-tuning LLaMA models on self-instruct and conversational data, respectively, demonstrating substantial gains in zero-shot task performance. Recent frameworks such as LLaMA-Factory \cite{zheng2024llamafactory} have further streamlined the fine-tuning process by providing unified interfaces for efficient training with parameter-efficient methods and distributed optimization.

In multilingual settings, prior work has shown that fine-tuning with a carefully balanced multilingual mixture preserves cross-lingual transfer while specializing to target languages \cite{xia2023sheared}. Our approach follows this principle by including a modest Chinese component in the supervised fine-tuning stage, which serves as an anchor to maintain the model's multilingual capabilities while primarily focusing on Tibetan task specialization.

\section{Method}
\begin{figure*}[t]
  \centering
  \includegraphics[width=\linewidth]{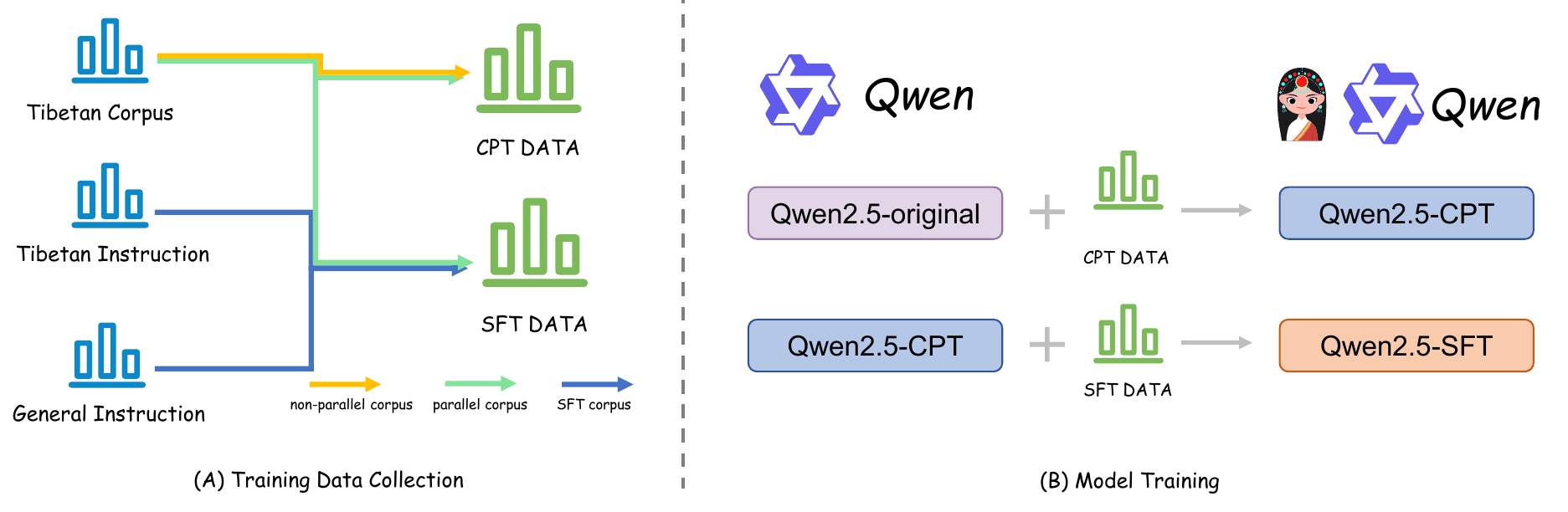}
  \caption{Two-stage training pipeline and data composition. Continual pretraining (CPT) uses Tibetan monolingual text derived from parallel corpora and non-parallel sources; supervised fine-tuning (SFT) uses an instruction mixture (BO→BO, CN→BO, EN→BO) with a small general Chinese component for multilingual anchoring.}
  \label{fig:pipeline}
\end{figure*}

\subsection{Overview}
Our approach employs a two-stage training pipeline to adapt Qwen2.5-3B \cite{hui2024qwen2} to Tibetan under low-resource constraints. The first stage, Continual Pretraining (CPT), establishes linguistic grounding by exposing the model to large-scale Tibetan monolingual text. The second stage, Supervised Fine-Tuning (SFT), refines the model's capabilities for specific tasks such as instruction following and translation by training on task-annotated data with explicit supervision.

This separation of linguistic adaptation and task alignment reduces catastrophic forgetting and optimization interference between distributional adaptation and task supervision \cite{gururangan2020dont}. In the first stage, CPT adapts the model to the Tibetan text distribution, recalibrating embeddings and the prediction space; the second stage (SFT) then specializes instruction-following and translation with minimal drift. Qwen2.5-3B is pretrained on multilingual corpora and uses a tokenizer that already covers Tibetan script, which offers basic token coverage but not task competence. Consequently, CPT provides the bulk of Tibetan exposure, while SFT primarily consolidates behavior and task formatting.

\subsection{Continual Pretraining}
The objective of continual pretraining is to establish robust Tibetan linguistic grounding by adapting the base model's representations to the lexical, syntactic, and discourse properties of Tibetan. We assemble a corpus of 200,000 Tibetan-only texts drawn from two primary sources: the CUTE dataset\cite{zhuang2025cute} (parallel Tibetan portions and non-parallel Tibetan documents) and the tibetan-mix collection. These texts span diverse genres including news, literature, and religious texts, providing broad coverage of Tibetan linguistic variation.

To expose the model to document-level context during CPT, we set the maximum context length to 8{,}192 tokens. This choice allows the model to learn dependencies beyond the sentence level and reduces truncation artifacts on long Tibetan texts. CPT is run for one pass over the corpus under the standard left-to-right causal language modeling objective (next-token prediction), without task-specific supervision.

The use of monolingual Tibetan data in CPT ensures that the model's internal representations are re-calibrated toward Tibetan token distributions and linguistic patterns, without the confounding influence of task-specific supervision. This stage can be viewed as domain-adaptive pretraining \cite{gururangan2020dont}, where the "domain" is the Tibetan language itself.

\subsection{Supervised Fine-Tuning}
Following continual pretraining, we apply supervised fine-tuning to align the model with specific tasks and instruction-following behaviors. The SFT corpus consists of 50,000 examples with a carefully designed 80/20 mixture: 80\% Tibetan-focused tasks and 20\% Chinese general instructions.

The Tibetan component includes three task categories:
\begin{itemize}
    \item \textbf{Tibetan instruction following (BO→BO):} Question-answering and instruction-completion tasks where both input and output are in Tibetan, promoting fluency and task adherence in the target language.
    \item \textbf{Chinese-to-Tibetan translation (CN→BO):} Parallel sentence pairs from CUTE and tibetan-mix, formatted as translation instructions to teach the model explicit translation behavior.
    \item \textbf{English-to-Tibetan translation (EN→BO):} A smaller set of English-Tibetan pairs to enable cross-lingual transfer and multilingual translation capabilities.
\end{itemize}

The 20\% Chinese portion serves as a multilingual anchor that maintains the model's existing Chinese competence and mitigates forgetting of non-Tibetan knowledge acquired during base pretraining. Training is conducted for two epochs with a maximum sequence length of 4,096 tokens. All examples are formatted as instruction-response pairs, enabling the model to learn task conditioning and response generation in a unified framework. Translation prompts follow standardized templates (e.g., "Translate the following Chinese text to Tibetan:") to ensure consistent task framing.

\section{Experiments}

\subsection{Implementation Details}
We implement our two-stage training pipeline using LLaMA-Factory \cite{zheng2024llamafactory}, an efficient fine-tuning framework that supports distributed training and memory optimization. Training is conducted on 8×NVIDIA H20 GPUs with BF16 mixed precision, leveraging DeepSpeed ZeRO-2 \cite{rasley2020deepspeed} for gradient sharding and memory efficiency.

For continual pretraining, we adopt an effective batch size of 128, a learning rate of 5e-5, and train for one epoch over the 200K Tibetan corpus with a maximum sequence length of 8,192 tokens. For supervised fine-tuning, we use an effective batch size of 128, reduce the learning rate to 1e-5 to preserve CPT-acquired knowledge, and train for two epochs over the 50K instruction corpus with a maximum sequence length of 4,096 tokens. Both stages use the AdamW optimizer with cosine learning rate scheduling and warmup.

Evaluation is conducted on three checkpoints: the base Qwen2.5-3B model, the CPT checkpoint after one epoch, and the final SFT checkpoint at epoch 2. For perplexity measurement, we fix the maximum context length to 4,096 tokens across all stages to ensure fair comparison. For translation evaluation, we use beam search with 4 beams, a temperature of 0.7, and generate up to 256 new tokens. Translation quality is assessed using BLEU and chrF metrics on held-out test sets filtered to match the sequence length constraints of each stage.

\subsection{Quantitative Analysis}

\begin{table}[t]
\centering
\small
\begin{tabular}{lccc}
\toprule
\textbf{Metric} & \textbf{Base} & \textbf{CPT} & \textbf{SFT} \\
\midrule
Perplexity & 2.98 & 1.61 & \textbf{1.54} \\
\midrule
\multicolumn{4}{l}{\textit{Chinese→Tibetan Translation}} \\
BLEU & 0.046 & 0.121 & \textbf{0.261} \\
chrF & 2.2 & 4.3 & \textbf{6.6} \\
\midrule
\multicolumn{4}{l}{\textit{English→Tibetan Translation}} \\
BLEU & 0.038 & 0.095 & \textbf{0.186} \\
chrF & 1.9 & 3.7 & \textbf{5.4} \\
\bottomrule
\end{tabular}
\caption{Performance progression across training stages. Perplexity decreases monotonically, while translation quality (BLEU and chrF) improves consistently from Base through CPT to SFT. The largest gains occur after SFT for Chinese→Tibetan due to explicit translation supervision, with secondary improvements on English→Tibetan via cross-lingual transfer.}
\label{tab:results}
\end{table}

Table \ref{tab:results} summarizes the quantitative improvements across training stages. Perplexity exhibits a consistent downward trend from the base model (2.98) through CPT (1.61) to SFT (1.54), indicating progressive improvements in Tibetan language modeling without regression. This monotonic decrease demonstrates that CPT successfully establishes Tibetan linguistic grounding, and SFT further refines this foundation without disrupting the language model's core capabilities.

Translation quality shows substantial gains across both language pairs. For Chinese→Tibetan, BLEU improves from 0.046 (Base) to 0.261 (SFT)—a 5.7× increase—while chrF rises from 2.2 to 6.6—a 3× improvement. The largest gains occur after SFT, reflecting the presence of explicit Chinese→Tibetan translation pairs in the supervised training data. English→Tibetan translation also improves significantly (BLEU: 0.038 → 0.186; chrF: 1.9 → 5.4), despite having fewer training examples, demonstrating effective cross-lingual transfer from the model's multilingual pretraining and the shared instruction-following framework.

\begin{figure}[t]
  \centering
  \includegraphics[width=0.48\textwidth]{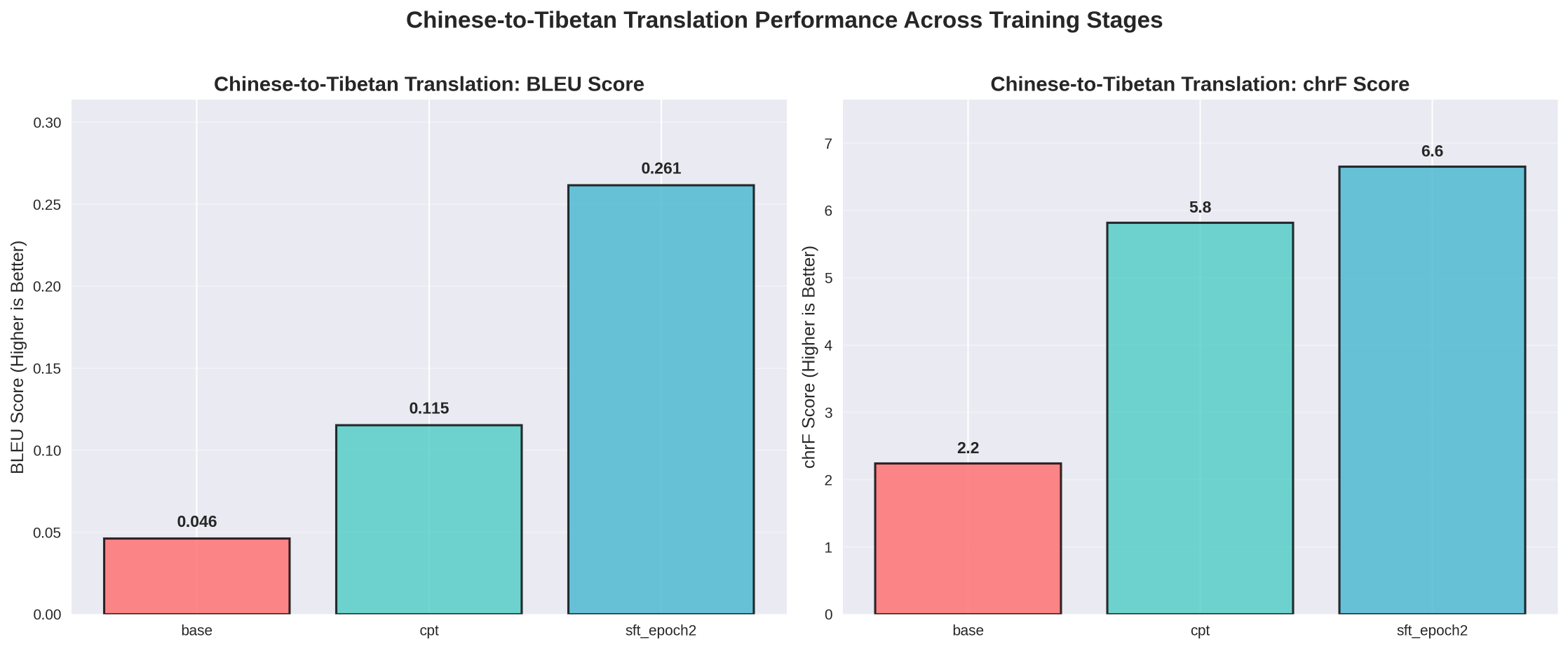}
  \caption{Translation quality progression across stages. Chinese→Tibetan exhibits the largest improvements due to explicit supervision in SFT, while English→Tibetan benefits from cross-lingual transfer.}
  \label{fig:trans_metrics}
\end{figure}

\begin{figure}[t]
  \centering
  \includegraphics[width=0.48\textwidth]{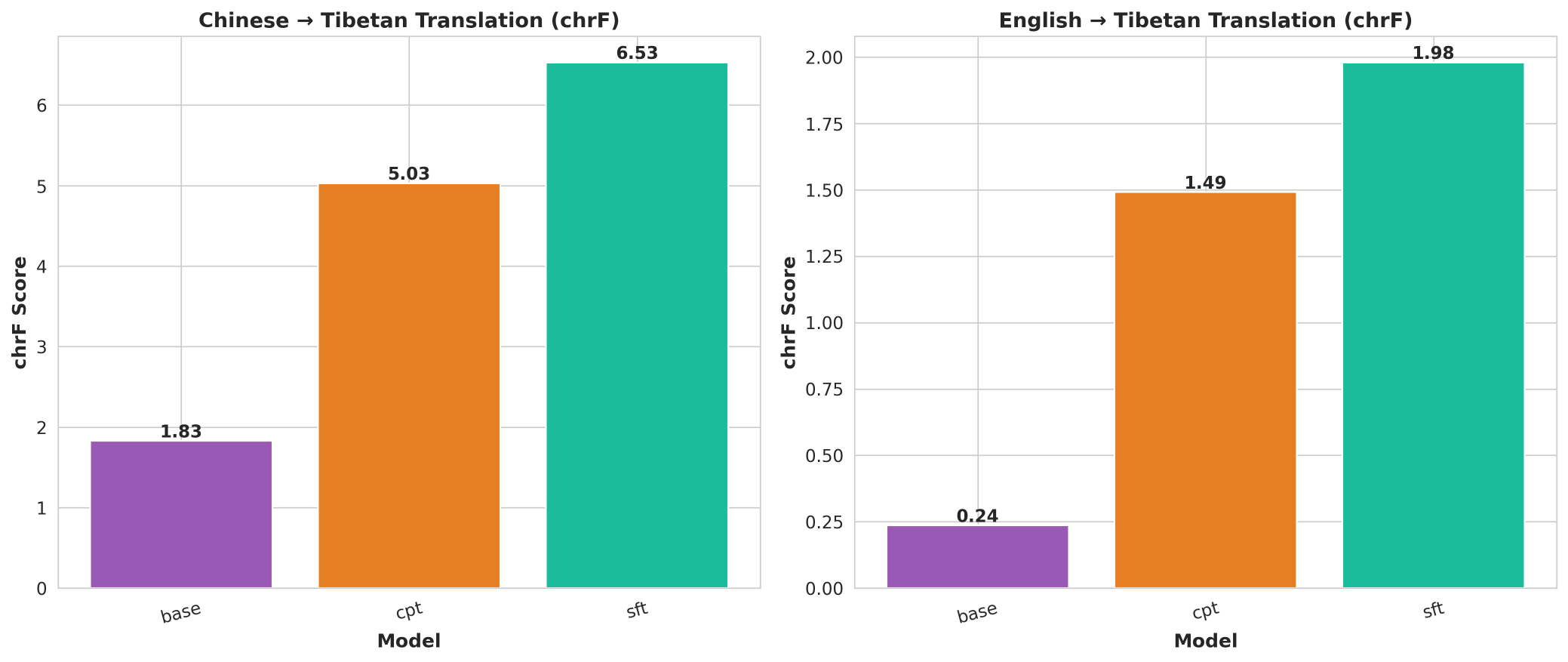}
  \caption{chrF score comparison highlighting consistent gains from Base to SFT for both translation directions.}
  \label{fig:chrf}
\end{figure}

Figures \ref{fig:trans_metrics} and \ref{fig:chrf} visualize the translation quality improvements. Both metrics demonstrate clear monotonic gains across stages, with CPT establishing emergent translation capabilities and SFT consolidating these into robust task performance. Notably, the absence of regression across any metric suggests that the two-stage approach successfully balances specialization with knowledge retention.

\subsection{Weight Change Analysis}
We study how parameters evolve across stages to understand \emph{where} and \emph{how} the model adapts. For each transformer block and for the embedding/output matrices (435 components in total), we compute layer-wise L2 norms of parameter deltas between checkpoints: $\Delta_{\text{Base}\rightarrow\text{CPT}}$, $\Delta_{\text{Base}\rightarrow\text{SFT}}$, and $\Delta_{\text{CPT}\rightarrow\text{SFT}}$. This gives a stage-aware view of update magnitude and locality, and lets us test whether SFT builds on, rather than overwrites, CPT-induced changes.

\paragraph{Global Statistics (descriptive)}
Table~\ref{tab:weight_stats} reports summary magnitudes. Because CPT and SFT differ in data and training budgets, these numbers are \emph{descriptive} and not sufficient to diagnose consolidation on their own. Relative to Base, CPT induces substantial movement (mean 2.29; max 28.6). SFT shows a similar scale vs.\ Base, while the incremental $\Delta_{\text{CPT}\rightarrow\text{SFT}}$ is smaller on average (mean 0.358; max 4.07). We therefore use these magnitudes as context and turn to localization and correlation evidence next.

\begin{table}[t]
\centering
\small
\begin{tabular}{lccc}
\toprule
\textbf{Comparison} & \textbf{Mean} & \textbf{Median} & \textbf{Max} \\
\midrule
CPT vs Base & 2.29 & 0.891 & 28.6 \\
SFT vs Base & 2.33 & 0.907 & 30.1 \\
SFT vs CPT & 0.358 & 0.152 & 4.07 \\
\bottomrule
\end{tabular}
\caption{Layer-wise L2 norms of parameter deltas (Base$ \leftrightarrow $CPT, Base$ \leftrightarrow $SFT, CPT$ \leftrightarrow $SFT). Values are not normalized for training budgets and are used as descriptive context alongside localization and correlation analyses.}
\label{tab:weight_stats}
\end{table}

\begin{figure}[t]
  \centering
  \includegraphics[width=0.48\textwidth]{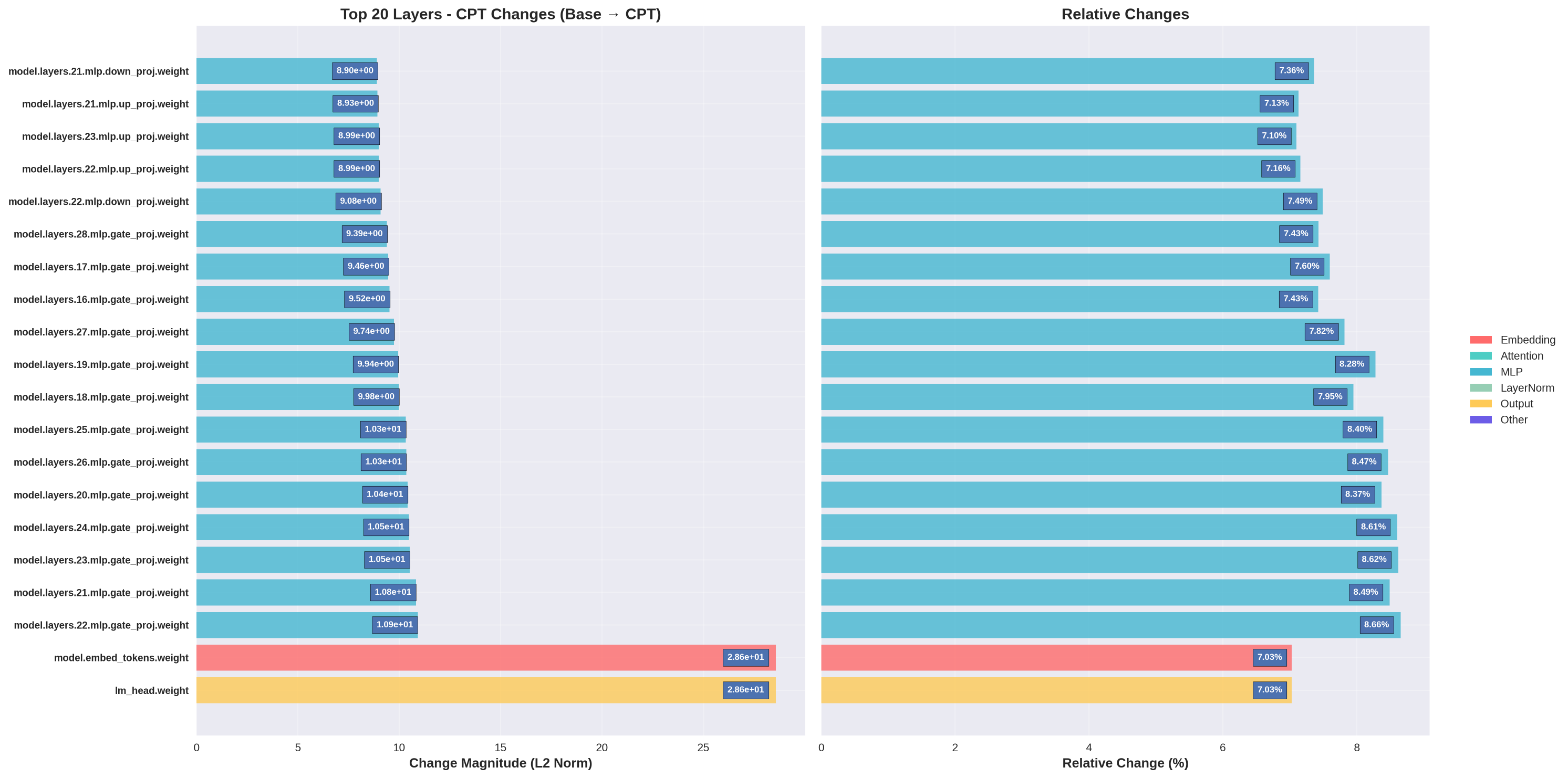}
  \caption{Top layer changes during CPT. Embeddings, output head, and mid-late MLP gate projections (layers 21-23) dominate.}
  \label{fig:cpt_changes}
\end{figure}

\begin{figure}[t]
  \centering
  \includegraphics[width=0.48\textwidth]{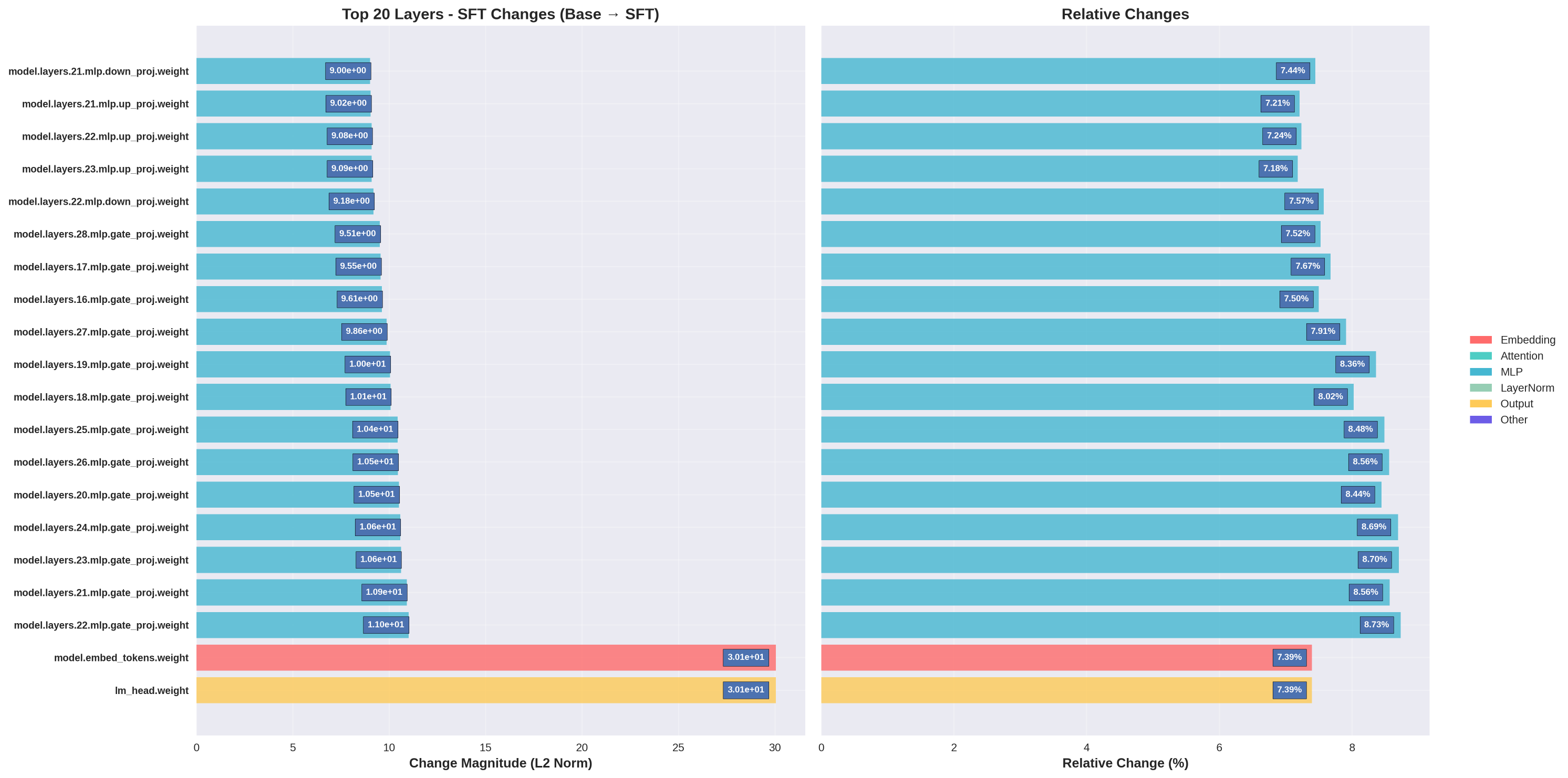}
  \caption{Top layer changes during SFT. Similar loci to CPT, with concentrated updates in embeddings, lm\_head, and layers 24-26.}
  \label{fig:sft_changes}
\end{figure}

\paragraph{Where do changes happen?}
Figures~\ref{fig:cpt_changes} and \ref{fig:sft_changes} show that updates are highly localized. The largest shifts occur in \texttt{model.embed\_tokens.weight} and \texttt{lm\_head.weight}, followed by mid–late MLP gate projections. This pattern is consistent across stages and indicates (i) re-anchoring of token representations and the logit space toward Tibetan distributions, and (ii) specialization of deep MLP transformations that support fluent Tibetan generation.

\begin{figure}[t]
  \centering
  \includegraphics[width=0.45\textwidth]{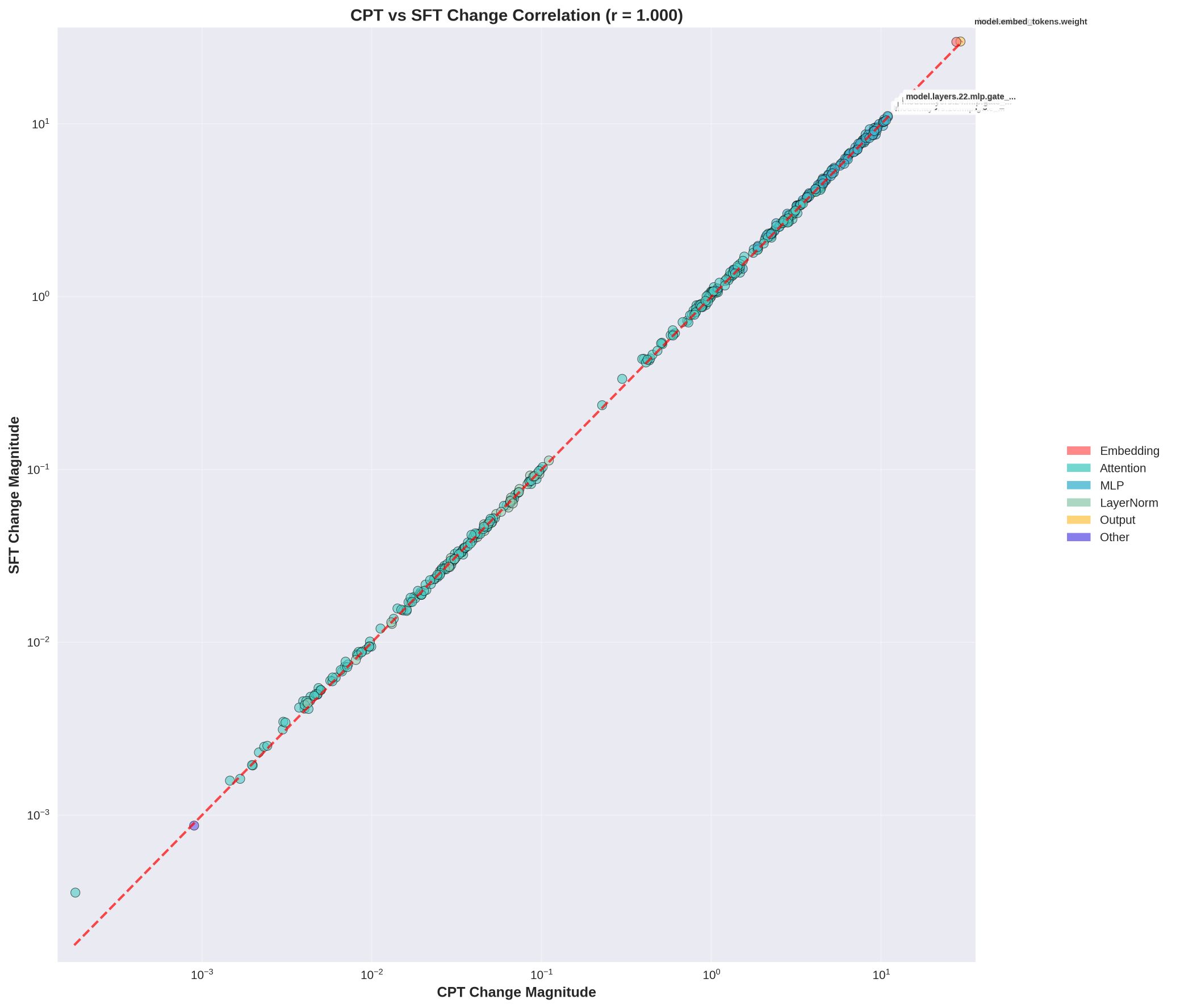}
  \caption{Correlation between CPT and SFT weight changes ($r \approx 1.0$). SFT consolidates CPT adaptations rather than introducing orthogonal shifts.}
  \label{fig:correlation}
\end{figure}

\begin{figure}[t]
  \centering
  \includegraphics[width=0.45\textwidth]{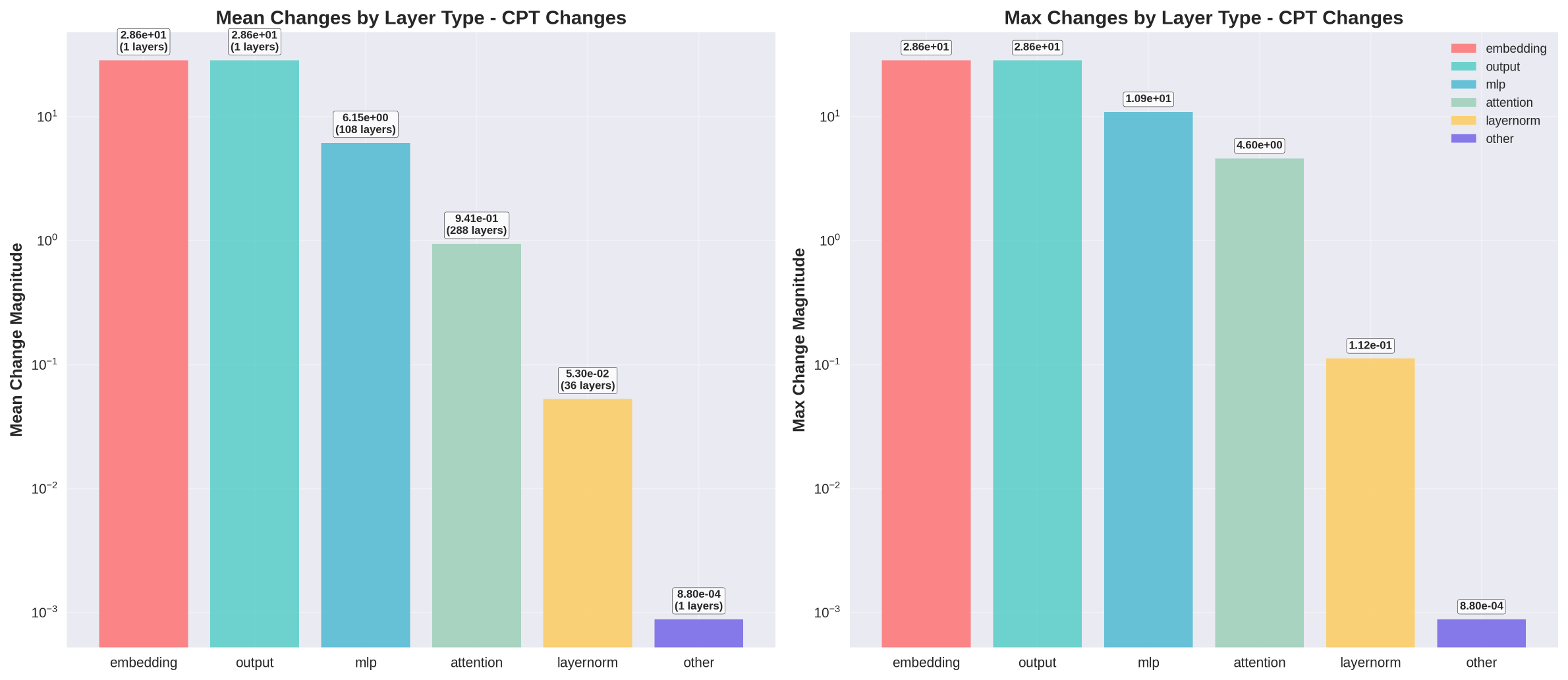}
  \caption{CPT weight changes by layer type. Embeddings and lm\_head dominate, with secondary MLP updates.}
  \label{fig:cpt_types}
\end{figure}

\begin{figure}[t]
  \centering
  \includegraphics[width=0.45\textwidth]{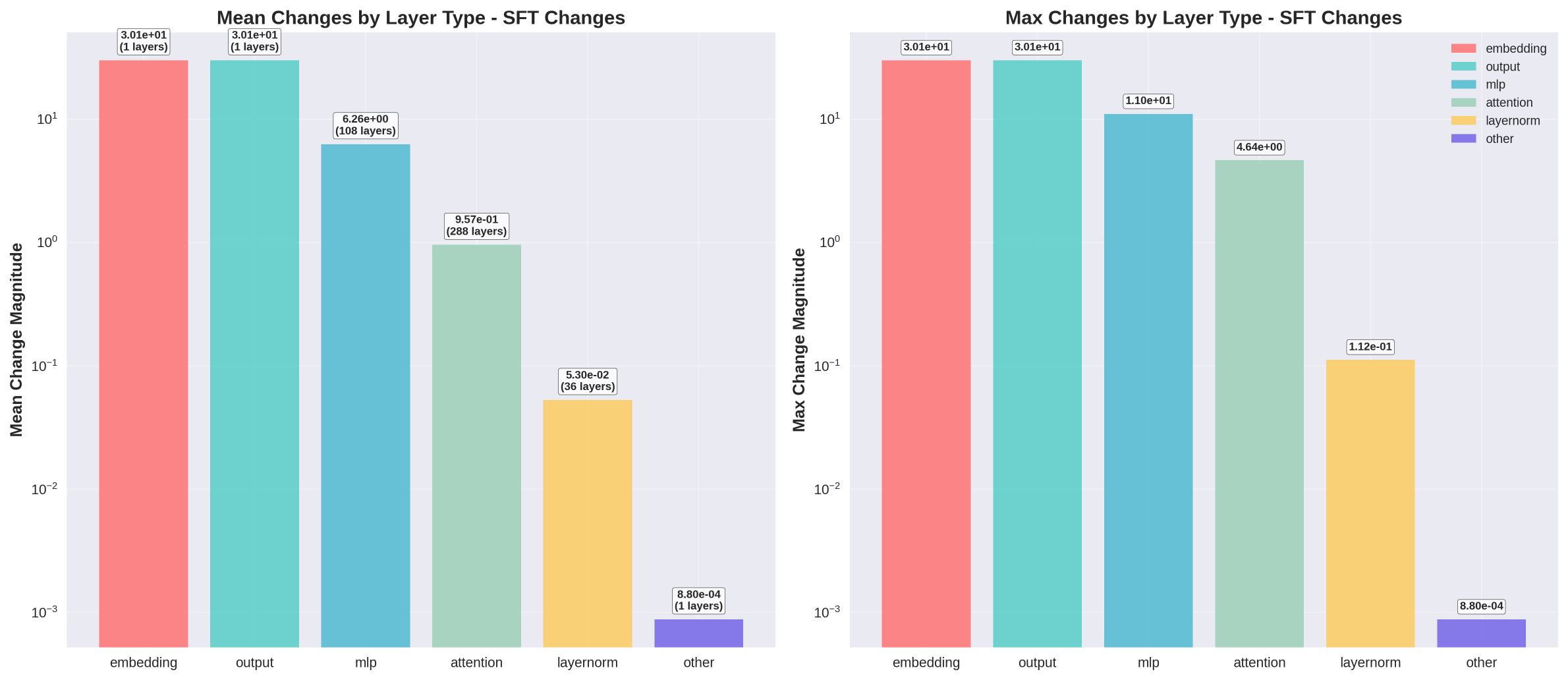}
  \caption{SFT weight changes by layer type. Pattern consistent with CPT, reinforcing the localization hypothesis.}
  \label{fig:sft_types}
\end{figure}

\paragraph{Are SFT updates aligned with CPT?}
Alignment, rather than magnitude alone, is the critical test for consolidation. The CPT–SFT correlation plot in Fig.~\ref{fig:correlation} shows near-linear alignment between per-layer deltas ($r\!\approx\!1.0$), indicating that SFT operates largely in the subspace established by CPT rather than redirecting updates to new regions of the network. The layer-type breakdowns in Figs.~\ref{fig:cpt_types} and \ref{fig:sft_types} further confirm that both stages emphasize embeddings/output heads with secondary MLP updates.

\paragraph{Takeaways}
Putting these signals together: (1) CPT performs the heavy lifting by re-anchoring embeddings and the output head and by specializing mid–late MLPs; (2) SFT concentrates updates in the same loci with smaller incremental norms; and (3) the near-unity correlation across layers demonstrates that SFT consolidates CPT-induced representations instead of rewriting them.

\paragraph{Interpretation}
The localization of weight changes to embeddings, output heads, and mid-late MLPs suggests three key mechanisms: (1) \textit{embedding re-anchoring}, where token representations are re-gauged toward Tibetan lexical semantics; (2) \textit{logit space alignment}, where the output projection learns to prioritize Tibetan tokens during generation; and (3) \textit{deep MLP specialization}, where mid-to-late layers encode domain-specific transformations for Tibetan discourse structure. The high correlation between CPT and SFT changes ($r \approx 1.0$) indicates that SFT primarily sharpens and consolidates the linguistic manifold established during CPT, rather than introducing fundamentally new adaptations. This consolidation mechanism may explain the absence of catastrophic forgetting and the monotonic performance improvements observed in our quantitative results.

\section{Conclusion}

This work presents a systematic two-stage adaptation of Qwen2.5-3B to Tibetan, combining continual pretraining for linguistic grounding with supervised fine-tuning for task specialization. Empirical evaluation demonstrates consistent improvements: perplexity decreases from 2.98 to 1.54, and Chinese→Tibetan translation quality increases not only by over 5× in BLEU (0.046→0.261), but 3× in chrF (2.2→6.6). Layer-wise weight analysis across 435 layers reveals that adaptation concentrates primarily in token embeddings and the output projection, with secondary updates in mid-to-late MLP layers; the near-unity correlation between CPT and SFT weight changes ($r \approx 1.0$) indicates that supervised fine-tuning consolidates rather than overwrites the linguistic manifold established during continual pretraining. This study demonstrates a practical and reproducible framework for extending multilingual foundation models to low-resource languages, contributing to more equitable access to LLM capabilities for under-resourced linguistic communities.

\bibliography{custom}

\begin{thebibliography}{22}
\providecommand{\natexlab}[1]{#1}

\bibitem[{Achiam et~al.(2023)Achiam, Adler, Agarwal, Ahmad, Akkaya, Aleman, Almeida, Altenschmidt, Altman, Anadkat et~al.}]{achiam2023gpt}
Josh Achiam, Steven Adler, Sandhini Agarwal, Lama Ahmad, Ilge Akkaya, Florencia~Leoni Aleman, Diogo Almeida, Janko Altenschmidt, Sam Altman, Shyamal Anadkat, and 1 others. 2023.
\newblock Gpt-4 technical report.
\newblock \emph{arXiv preprint arXiv:2303.08774}.

\bibitem[{Bai et~al.(2023)Bai, Bai, Chu, Cui, Dang, Deng, Fan, Ge, Han, Huang, Hui, Ji, Li, Lin, Lin, Liu, Liu, Lu, Lu, Ma, Men, Ren, Ren, Tan, Tan, Tu, Wang, Wang, Wang, Wu, Xu, Xu, Yang, Yang, Yang, Yang, Yao, Yu, Yuan, Yuan, Zhang, Zhang, Zhang, Zhang, Zhou, Zhou, Zhou, and Zhu}]{bai2023qwentechnicalreport}
Jinze Bai, Shuai Bai, Yunfei Chu, Zeyu Cui, Kai Dang, Xiaodong Deng, Yang Fan, Wenbin Ge, Yu~Han, Fei Huang, Binyuan Hui, Luo Ji, Mei Li, Junyang Lin, Runji Lin, Dayiheng Liu, Gao Liu, Chengqiang Lu, Keming Lu, and 29 others. 2023.
\newblock \href {https://arxiv.org/abs/2309.16609} {Qwen technical report}.
\newblock \emph{Preprint}, arXiv:2309.16609.

\bibitem[{Chiang et~al.(2023)Chiang, Li, Lin, Sheng, Wu, Zhang, Zheng, Zhuang, Zhuang, Gonzalez et~al.}]{chiang2023vicuna}
Wei-Lin Chiang, Zhuohan Li, Zi~Lin, Ying Sheng, Zhanghao Wu, Hao Zhang, Lianmin Zheng, Siyuan Zhuang, Yonghao Zhuang, Joseph~E Gonzalez, and 1 others. 2023.
\newblock Vicuna: An open-source chatbot impressing gpt-4 with 90\%* chatgpt quality.
\newblock \emph{See https://vicuna. lmsys. org (accessed 14 April 2023)}.

\bibitem[{Conneau et~al.(2020)Conneau, Khandelwal, Goyal, Chaudhary, Wenzek, Guzm{\'a}n, Grave, Ott, Zettlemoyer, and Stoyanov}]{conneau-etal-2020-unsupervised}
Alexis Conneau, Kartikay Khandelwal, Naman Goyal, Vishrav Chaudhary, Guillaume Wenzek, Francisco Guzm{\'a}n, Edouard Grave, Myle Ott, Luke Zettlemoyer, and Veselin Stoyanov. 2020.
\newblock \href {https://doi.org/10.18653/v1/2020.acl-main.747} {Unsupervised cross-lingual representation learning at scale}.
\newblock In \emph{Proceedings of the 58th Annual Meeting of the Association for Computational Linguistics}, pages 8440--8451, Online. Association for Computational Linguistics.

\bibitem[{Costa-Juss{\`a} et~al.(2022)Costa-Juss{\`a}, Cross, {\c{C}}elebi, Elbayad, Heafield, Heffernan, Kalbassi, Lam, Licht, Maillard et~al.}]{costa2022no}
Marta~R Costa-Juss{\`a}, James Cross, Onur {\c{C}}elebi, Maha Elbayad, Kenneth Heafield, Kevin Heffernan, Elahe Kalbassi, Janice Lam, Daniel Licht, Jean Maillard, and 1 others. 2022.
\newblock No language left behind: Scaling human-centered machine translation.
\newblock \emph{arXiv preprint arXiv:2207.04672}.

\bibitem[{Devlin et~al.(2019)Devlin, Chang, Lee, and Toutanova}]{devlin2019bert}
Jacob Devlin, Ming-Wei Chang, Kenton Lee, and Kristina Toutanova. 2019.
\newblock Bert: Pre-training of deep bidirectional transformers for language understanding.
\newblock In \emph{Proceedings of the 2019 conference of the North American chapter of the association for computational linguistics: human language technologies, volume 1 (long and short papers)}, pages 4171--4186.

\bibitem[{Gururangan et~al.(2020)Gururangan, Marasovi{\'c}, Swayamdipta, Lo, Beltagy, Downey, and Smith}]{gururangan2020dont}
Suchin Gururangan, Ana Marasovi{\'c}, Swabha Swayamdipta, Kyle Lo, Iz~Beltagy, Doug Downey, and Noah~A Smith. 2020.
\newblock Don't stop pretraining: Adapt language models to domains and tasks.
\newblock \emph{arXiv preprint arXiv:2004.10964}.

\bibitem[{He et~al.(2023)He, Gesang, Qun, and Nyima}]{10455584}
Chenghao He, Quzong Gesang, Nuo Qun, and Tashi Nyima. 2023.
\newblock \href {https://doi.org/10.1109/DSInS60115.2023.10455584} {Research on tibetan-chinese neural machine translation based on gru}.
\newblock In \emph{2023 3rd International Conference on Digital Society and Intelligent Systems (DSInS)}, pages 213--216.

\bibitem[{Huang et~al.(2024)Huang, Zhang, Li, Chen, and Wang}]{huang2024tibetan}
Ying Huang, Jian Zhang, Wei Li, Ming Chen, and Xiao Wang. 2024.
\newblock Tibetan language and ai: A comprehensive survey of resources, methods and challenges.
\newblock \emph{arXiv preprint arXiv:2510.19144}.

\bibitem[{Hui et~al.(2024)Hui, Yang, Cui, Yang, Liu, Zhang, Liu, Zhang, Yu, Lu et~al.}]{hui2024qwen2}
Binyuan Hui, Jian Yang, Zeyu Cui, Jiaxi Yang, Dayiheng Liu, Lei Zhang, Tianyu Liu, Jiajun Zhang, Bowen Yu, Keming Lu, and 1 others. 2024.
\newblock Qwen2. 5-coder technical report.
\newblock \emph{arXiv preprint arXiv:2409.12186}.

\bibitem[{Joshi et~al.(2020)Joshi, Santy, Budhiraja, Bali, and Choudhury}]{joshi2020state}
Pratik Joshi, Sebastin Santy, Amar Budhiraja, Kalika Bali, and Monojit Choudhury. 2020.
\newblock The state and fate of linguistic diversity and inclusion in the nlp world.
\newblock \emph{arXiv preprint arXiv:2004.09095}.

\bibitem[{Kyi et~al.(2025)Kyi, Gyal, Tashi, and Duojie}]{11064426}
Lhamao Kyi, DungSkyabsr Gyal, Nyima Tashi, and Renzeng Duojie. 2025.
\newblock \href {https://doi.org/10.1109/NNICE64954.2025.11064426} {Construction of a parallel corpus of tibetan dialect}.
\newblock In \emph{2025 5th International Conference on Neural Networks, Information and Communication Engineering (NNICE)}, pages 685--689.

\bibitem[{Liu et~al.(2020)Liu, Gu, Goyal, Li, Edunov, Ghazvininejad, Lewis, and Zettlemoyer}]{liu2020multilingual}
Yinhan Liu, Jiatao Gu, Naman Goyal, Xian Li, Sergey Edunov, Marjan Ghazvininejad, Mike Lewis, and Luke Zettlemoyer. 2020.
\newblock Multilingual denoising pre-training for neural machine translation.
\newblock \emph{Transactions of the Association for Computational Linguistics}, 8:726--742.

\bibitem[{Rasley et~al.(2020)Rasley, Rajbhandari, Ruwase, and He}]{rasley2020deepspeed}
Jeff Rasley, Samyam Rajbhandari, Olatunji Ruwase, and Yuxiong He. 2020.
\newblock Deepspeed: System optimizations enable training deep learning models with over 100 billion parameters.
\newblock In \emph{Proceedings of the 26th ACM SIGKDD international conference on knowledge discovery \& data mining}, pages 3505--3506.

\bibitem[{Taori et~al.(2023)Taori, Gulrajani, Zhang, Dubois, Li, Guestrin, Liang, and Hashimoto}]{taori2023alpaca}
Rohan Taori, Ishaan Gulrajani, Tianyi Zhang, Yann Dubois, Xuechen Li, Carlos Guestrin, Percy Liang, and Tatsunori~B Hashimoto. 2023.
\newblock Alpaca: A strong, replicable instruction-following model.
\newblock \emph{Stanford Center for Research on Foundation Models}.

\bibitem[{Touvron et~al.(2023)Touvron, Lavril, Izacard, Martinet, Lachaux, Lacroix, Rozi{\`e}re, Goyal, Hambro, Azhar et~al.}]{touvron2023llama}
Hugo Touvron, Thibaut Lavril, Gautier Izacard, Xavier Martinet, Marie-Anne Lachaux, Timoth{\'e}e Lacroix, Baptiste Rozi{\`e}re, Naman Goyal, Eric Hambro, Faisal Azhar, and 1 others. 2023.
\newblock Llama: Open and efficient foundation language models.
\newblock \emph{arXiv preprint arXiv:2302.13971}.

\bibitem[{Wang et~al.(2024)Wang, Li, Nyima, and Zhou}]{wang2024tllama}
Qun Wang, Xiaobing Li, Tashi Nyima, and Yiwen Zhou. 2024.
\newblock \href {https://doi.org/10.1007/s40747-024-01641-7} {T-llama: A tibetan large language model based on llama2}.
\newblock \emph{Complex \& Intelligent Systems}.

\bibitem[{Wang et~al.(2025)Wang, Zhang, Li, and Chen}]{wang2025tlue}
Xiaolin Wang, Lei Zhang, Yun Li, and Bo~Chen. 2025.
\newblock Tlue: A tibetan language understanding evaluation benchmark.
\newblock \emph{arXiv preprint arXiv:2503.12051}.

\bibitem[{Xia et~al.(2023)Xia, Gao, Zeng, and Chen}]{xia2023sheared}
Mengzhou Xia, Tianyu Gao, Zhiyuan Zeng, and Danqi Chen. 2023.
\newblock Sheared llama: Accelerating language model pre-training via structured pruning.
\newblock \emph{arXiv preprint arXiv:2310.06694}.

\bibitem[{Zheng et~al.(2024)Zheng, Zhang, Zhang, Ye, Luo, and Ma}]{zheng2024llamafactory}
Yaowei Zheng, Richong Zhang, Junhao Zhang, Yanhan Ye, Zheyan Luo, and Yongqiang Ma. 2024.
\newblock Llama-factory: Unified efficient fine-tuning of 100+ language models.

\bibitem[{Zhou et~al.(2023)Zhou, Li, Wang, and Nyima}]{zhou2023peftt}
Yiwen Zhou, Xiaobing Li, Qun Wang, and Tashi Nyima. 2023.
\newblock Peftt: Parameter-efficient fine-tuning for low-resource tibetan pre-trained language models.
\newblock \emph{arXiv preprint arXiv:2309.12109}.

\bibitem[{Zhuang and Sun(2025)}]{zhuang2025cute}
Wenhao Zhuang and Yuan Sun. 2025.
\newblock Cute: A multilingual dataset for enhancing cross-lingual knowledge transfer in low-resource languages.
\newblock In \emph{Proceedings of the 31st International Conference on Computational Linguistics}, pages 10037--10046.

\end{thebibliography}

\end{document}